\documentclass[journal]{IEEEtran}

\usepackage{amsmath}
\usepackage{algorithmic}
\usepackage{algorithm}
\usepackage{array}

\usepackage[font=footnotesize,labelfont=rm,textfont=rm,caption=false]{subfig}

\usepackage{multirow}
\usepackage{textcomp}
\usepackage{stfloats}
\usepackage{url}
\usepackage{verbatim}
\usepackage{graphicx}
\usepackage{booktabs}
\usepackage{amssymb}
\usepackage{amsfonts}

\usepackage{cite}
\begin{document}

\title{ A Perception CNN for Facial Expression Recognition  }

\author{
Chunwei Tian,~\IEEEmembership{Member,~IEEE,} Jingyuan Xie, Lingjun Li, Wangmeng Zuo,~\IEEEmembership{Senior Member,~IEEE, }
\\ Yanning Zhang,~\IEEEmembership{Fellow,~IEEE,} and David Zhang,~\IEEEmembership{Life Fellow,~IEEE}

\thanks{
    This work was supported in part by National Natural Science Foundation
    of China under Grant 62201468 and 62576123. \textit{(Corresponding author: Lingjun Li, Yanning Zhang)}
}

\thanks{
Chunwei Tian and Wangmeng Zuo are with the School of Computer Science and Technology, Harbin Institute of Technology, Harbin, 150001, China (e-mail: chunweitian@hit.edu.cn, cswmzuo@gmail.com)
}
\thanks{
Jingyuan Xie is with the School of Software, Northwestern Polytechnical University, Xi’an, 710129, China. (e-mail: {xjyjason2001@mail.nwpu.edu.cn}).
}

\thanks{
    Lingjun Li is with the School of Software, Zhengzhou University of Light Industry, Zhengzhou, 450000, China.
    (e-mail: junllone2023@gmail.com).
}

\thanks{
    Yanning Zhang is with the School of Computer Science, Northwestern Polytechnical University, Xi’an, 710129, China.
    (e-mail: ynzhang@nwpu.edu.cn).
}

\thanks{
    David Zhang is with the School of Science and Engineering, Chinese University of Hong Kong, Shenzhen, 518172, China.
    (e-mail: davidzhang@cuhk.edu.cn).
}
}


\maketitle

\begin{abstract}
    Convolutional neural networks (CNNs) can automatically learn data patterns to express face images for facial expression recognition (FER). However, they may ignore effect of facial segmentation of FER. In this paper, we propose a perception CNN for FER as well as PCNN. Firstly, PCNN can use five parallel networks to simultaneously learn local facial features based on eyes, cheeks and mouth to realize the sensitive capture of the subtle changes in FER. Secondly, we utilize a multi-domain interaction mechanism to register and fuse between local sense organ features and global facial structural features to better express face images for FER. Finally, we design a two-phase loss function to restrict accuracy of obtained sense information and reconstructed face images to guarantee performance of obtained PCNN in FER. Experimental results show that our PCNN achieves superior results on several lab and real-world FER benchmarks: CK+, JAFFE, FER2013, FERPlus, RAF-DB and Occlusion and Pose Variant Dataset. Its code is available at https://github.com/hellloxiaotian/PCNN.
\end{abstract}


\section{Introduction}
\IEEEPARstart{I}{n} human social interaction, facial expressions are key non-verbal signals that convey emotions, intentions, and mental states. Thus, facial expression recognition techniques (FER) are important for security monitoring \cite{tao_frontal-centers_2022}, human-computer interaction \cite{khadidos_harnessing_2024}, mental health assessment \cite{sonawane_review_2021}, intelligent tutoring system \cite{jeong_drivers_2018}, etc. It aims to automatically recognize and interpret human facial expressions by analyzing and processing facial images \cite{li_deep_2022}. In the early stages of facial expression recognition research, traditional machine learning approaches primarily depend on appearance-based and geometric descriptors to extract handcrafted features from facial images to represent face images for FER \cite{corneanu_survey_2016}. For example, appearance features are designed to capture texture and surface information of facial images to express face images for FER \cite{ahonen_face_2006}. Geometric descriptors rely on the accurate localization of facial landmarks to exhibit physical interpretability, where mentioned features are constructed via measured distances, angles and relative positions between key points \cite{cootes_active_1995}. Furthermore, these geometric descriptors utilize geometric shape and motion changes of facial landmarks to represent expressions rather than facial textures for FER \cite{tian_recognizing_2001}. Local binary patterns utilized circular neighborhoods and uniform patterns to extract representative features to overcome gray-scale invariance and rotation invariance in FER \cite{ojala_multiresolution_2002}. To improve robustness of FER, histogram of oriented gradients can calculate gradient orientations within local image regions and organize them into block-based structures with contrast normalization to effectively obtain edge and shape features in FER \cite{dalal_histograms_2005}. To overcome effect of varying illumination and poses, Gabor wavelet filters can capture multi-scale and multi-orientation texture features to code facial expressions for better representing facial images for FER \cite{lyons_coding_1998}. Once these handcrafted features are extracted, some classifiers, i.e., support vector machines \cite{hearst_support_1998}, K-nearest neighbor \cite{cunningham_k-nearest_2021}, and decision trees \cite{costa_recent_2022} can be used to generate the result of expression recognition. Although these methods can recognize expressions, they are time-consuming and labor-intensive via manual feature extractions. That causes capacities of obtained features are inherently limited and these models cannot be suitable for complex scenes, i.e., subtle variations in FER \cite{li_deep_2022}.

With the rapid development of big data and graphics processing unit, deep learning techniques with deep network architectures can automatically learn features rather than relying on handcrafted features, thus becoming mainstream tools for facial expression recognition \cite{li_deep_2022}. Furthermore, flexible end-to-end CNN architectures, i.e., VGG \cite{simonyan_very_2014} and ResNet \cite{he_deep_2016} can meet requirements of FER of different scenes, i.e., varying expressions and occlusions. Liu et at. combined a CNN and random forest to address data dependency and improve accuracy of FER for occlusion \cite{liu_conditional_2018}. To improve robustness of FER, Mohan et al. \cite{mohan_facial_2020} used local gravitational force to extract edge details information of facial gradient and directions and exploited a deep network to extract geometry features, i.e., edge and curve. After fusion of two obtained features via a score fusion mechanism, performance of FER can be improved. Alternatively, Fei et al. \cite{fei_deep_2020} combined a CNN and linear data analysis to efficiently extract depth features to improve adaptive ability of an obtained classifier for FER.

\begin{figure}[!t]
    \centering
    \subfloat[][human face with occlusion]{\includegraphics[width=0.4\linewidth]{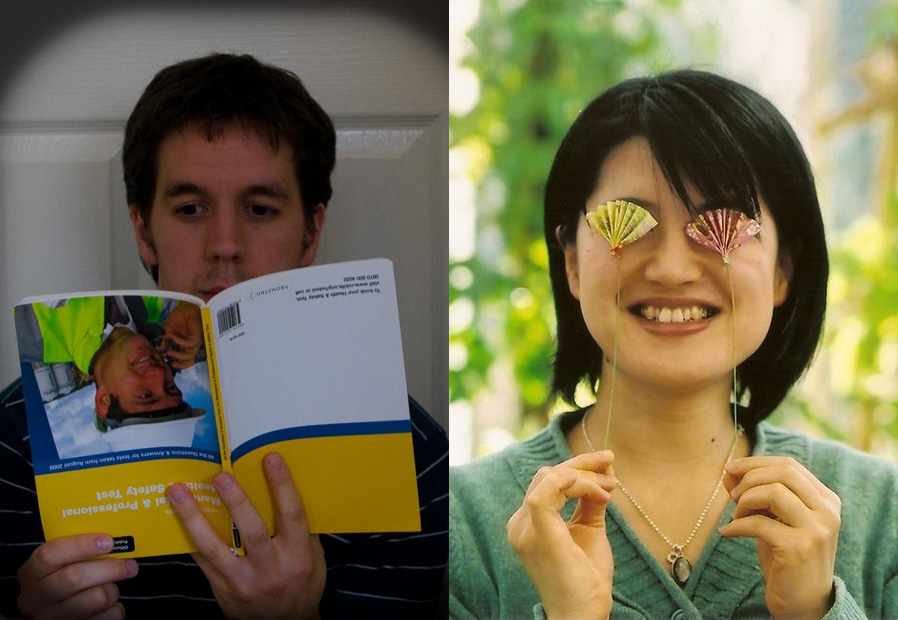}%
        \label{fig_occ_case}}
    \hfil
    \subfloat[][human face with pose changes]{\includegraphics[width=0.4\linewidth]{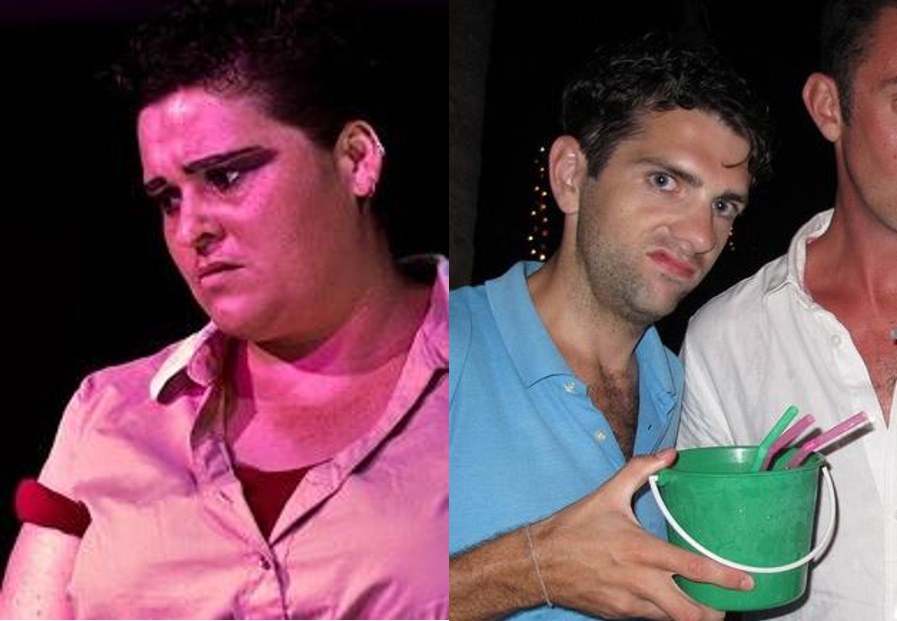}%
        \label{fig_pose_case}}
    \caption{The collected facial expression images in RAF-DB dataset (a) human face with occlusion (b) human face with pose changes.}
    \label{fig_sim}
\end{figure}

Although CNNs have achieved impressive results in ideal scenes, their performance often degrades in real-world applications, i.e., facial occlusion and varying head poses. In practical scenarios, facial regions may be partially obscured by external objects, i.e., masks, glasses, hands, and hair, or self-influenced by extreme head rotations, where these occlusions can result in critical local feature loss containing mouth, i.e., Fig. \ref{fig_occ_case}, eyes and eyebrow to decrease performance of FER. Varying head poses, i.e., non-frontal facial orientations containing yaw, pitch, roll axes can not only change visible shape and texture of facial features, but also disrupt their spatial relationships to effect extraction of effective features. As presented in Fig. \ref{fig_pose_case}, a face in a pronounced profile view may obscure one side of the mouth and eyebrow to affect accuracy of identifying asymmetric expressions, i.e., a smirk or a frown.

To mitigate these challenges, attention mechanisms have been referred to CNNs to extract more informative facial expression information for FER \cite{zhang_facial_2018}. For instance, Wang et al. \cite{k_wang_region_2020} developed region attention networks (RAN) via dynamically assigning weights to different facial regions to better capture subtle expression information for FER. Subsequently, Chen et al. \cite{chen_multi-relations_2023} adopted multi-scale feature fusion strategies by integrating local texture features and global structural information to enhance robustness against pose variations and occlusions. Alternatively, Zhao et al. \cite{zhao_geometry-aware_2021} employed a graph convolutional network to model structural relationships among facial components to further improve the resilience of a FER model. Additionally, data augmentation techniques based on generative models, i.e., GANs \cite{sun_discriminatively_2023} have been utilized to exploit unlabeled data and boost generalization performance in FER \cite{halawa_multi-task_2024}. Although these methods can perform well via extracting facial structural information, they ignore simultaneous use of five sense information and structural information to decrease effect of obtained classifiers in FER.

In this paper, we propose a Perception CNN (PCNN) via fully interacting structural information and perception information based sense organs for facial expression recognition. Firstly, PCNN utilizes a backbone network to extract global facial structural information and five parallel sub-networks to extract five sense information to obtain more accurate fine-grained expression change information. Secondly, PCNN exploits a two-phase multi-domain interaction mechanism to gather obtained fine-grained expression change information and obtained whole facial information for FER. The first phase utilizes a spatial transformer network to register obtained five local sense organ features. The second phase employs a residual learning operation to merge obtained global facial information and registered local sense organ information from the first phase to improve accuracy of FER. Finally, to prevent accuracy decrease of registration operation above, we design a two-phase loss function to ensure reliability of obtained sense information and verify performance of obtained PCNN via measuring differences between predicted result from reconstructed face images and given references. Its contributions can be summarized as follows.

(1) Using facial semantic information based on some sense organs guides deep networks to extract more salient detailed information for FER.

(2) A cross-domain interaction mechanism can better fuse reconstruction facial image based on sense organs and original facial images to better represent varying fine-gained expression to improve performance of FER.

(3) Facial semantic loss can guarantee accuracy of obtained local emotional information to further improve effect of our model for FER.

The remainder of this paper is structured as outlined below. Section II reviews traditional machine learning algorithms and deep learning algorithms for facial expression recognition. Section III describes PCNN in details. Section IV provides comprehensive experiments on six datasets of our proposed PCNN. Finally, we present the conclusion in Section V.

\section{RELATED WORKS}
In this section, we will review related works, i.e. traditional machine learning methods and deep learning methods for facial expression recognition.
\subsection{Traditional Machine Learning Methods for Facial Expression Recognition}
Facial Expression Recognition (FER) as an important branch of the field of computer vision, predominantly relied on traditional machine learning pipelines. The majority of these methods refer to manually designing feature extractors to train classification models. Most of these methods focus on developing and optimizing various handcrafted feature descriptors.

Early studies emphasized the effectiveness of basic texture and gradient features to represent facial appearance. For example, Lyons et al. \cite{lyons_coding_1998} applied Gabor wavelet filters to capture multi-scale and multi-directional texture characteristics. To obtain richer texture information of facial images, scholars extended grayscale texture information to color texture information to improve robustness of an obtained classifier of FER \cite{lajevardi_facial_2012}. That is, Lajevardi et al. \cite{lajevardi_facial_2012} proposed a tensor framework via perceptual color spaces to analyze value of color information to deal with FER of varying illumination. Subsequently, Ojala et al. \cite{ojala_multiresolution_2002} developed the Local Binary Pattern (LBP) via binary encoding obtained relations between neighbor and current pixels to achieve a computationally efficient descriptor for capturing more accurate local texture information to implement a robust person-independent FER algorithm \cite{shan_facial_2009}. Although LBP methods above are effective for FER with certain emotions, they still suffer from a challenge of FER of varying emotions. To overcome this drawback, LBP is extended to multi-domain spaces to extract more representative features for better expressing facial images for FER \cite{dalal_histograms_2005}, \cite{zhao_dynamic_2007}. Zhao et al. \cite{zhao_dynamic_2007} implemented local binary encoding in spatial-temporal domain to overcome effects of micro-expression for FER. Kotsia et al. \cite{kotsia_facial_2006} utilized geometric deformation of facial grid nodes to track muscle movements for FER. To improve robustness of FER, Li et al. \cite{li_simultaneous_2013} proposed a Dynamic Bayesian Network via constructing a probabilistic model of causal relationship between features and expressions to simultaneously track facial features and recognize expressions. These methods are suitable to certain scenes datsets, i.e. CK+ \cite{lucey_extended_2010} and MMI \cite{valstar_induced_2010}. However, due to manual feature extraction \cite{li_deep_2022}, specific pattern learning \cite{ghazi_comprehensive_2016} and loss of global facial information \cite{wang_feature_2013}, they may struggle in challenging to recognize facial expressions in real-world scenarios. To overcome these difficulties, feature fusion methods are developed. Wang et al. \cite{wang_feature_2013} fused two different feature descriptors to capture more comprehensive local features to improve the effect of FER. To further improve adaptive ability of an obtain facial classifier, we use a perception CNN via combining sense information and structural information to better represent facial emotions for enhancing robustness of our method for FER under complex scenes in this paper.
\subsection{Deep Learning methods for facial Expression Recognition}
Due to strong computational devices of GPU and data of large scale, convolutional neural networks (CNNs) reply on flexible end-to-end architectures containing VGGNet \cite{simonyan_very_2014} and ResNet \cite{he_deep_2016} can automatically learn patterns from data rather than handcrafted features to establish models of facial expression recognitions, which are more capable for FER in the real world \cite{zhang_facial_2018}. These methods can be summarized into two categories, holistic and region methods \cite{zhao_learning_2021}. Holistic methods directly utilized CNNs to act entire face images to learn comprehensive global feature representation to mine emotional information for FER. Also, early CNNs may have low learning efficiency in training to facilitate salient emotional information for FER. To address this problem, Xie et al. \cite{xie_adaptive_2021} used traditional handmade features as prior knowledge to guide a deep network to quickly train a classification model of FER. To deal with effect of lighting and varying poses on FER, Ni et al. \cite{ni_facial_2022} presented a cross-model network via fusing different features from RGB, depth, and thermal images to capture more texture information for FER. These methods can only use current facial features to analyze expressions rather than processing expression generation by muscle changes, which limits their good performance in FER.  To tackle this issue, Yu et al. \cite{yu_co-attentive_2022} introduced a multi-task network by learning the relationship between muscle changes and facial features from two sub-networks to fuse these features via a collaborative attention in FER. To obtain a more robust classifier of FER, Ruan et al. \cite{ruan_feature_2021} designed a two-stage network of decomposition and reconstruction via extracting common features of a certain type of expression and refining unique features of different expressions to obtain more refined facial expression features to improve generalization ability of the proposed FER algorithm. Although these holistic methods have significantly advanced global feature learning, they often overlook the effect of local facial region features, i.e. eyes and mouth, leading to decrease accuracy of distinguishing similar expressions.

Region methods are developed to directly address this mentioned limitation by explicitly modeling local facial features. A primary challenge of this method is that occlusions and pose variations can obscure these local critical regions to cause low recognition rate of FER. To handle this problem, Wang et al. \cite{k_wang_region_2020} presented a region attention network via dynamically assigning weights to different facial regions to focus on the most informative un-occluded areas for FER. To further improve practicality of an obtained FER algorithm, strengthening relationships between global and local information is presented for FER.  Chen et al. \cite{chen_multi-relations_2023} used a transformer to model internal relationships between local regions, fusion between global and local features and similarity relationship between different training samples to learn more comprehensive facial expression features for FER. Besides, Liu et al. \cite{liu_adaptive_2022} utilized an adaptive attention mechanism to guide a CNN to capture key texture details to capture more accurate emotional information. Differing from these methods, we propose a perception CNN via simultaneously using local sense organ information and global facial information to overcome effect of facial expression recognition with complex scenarios in this paper.

\begin{figure*}[!t]\centering
    \includegraphics[width=\textwidth]{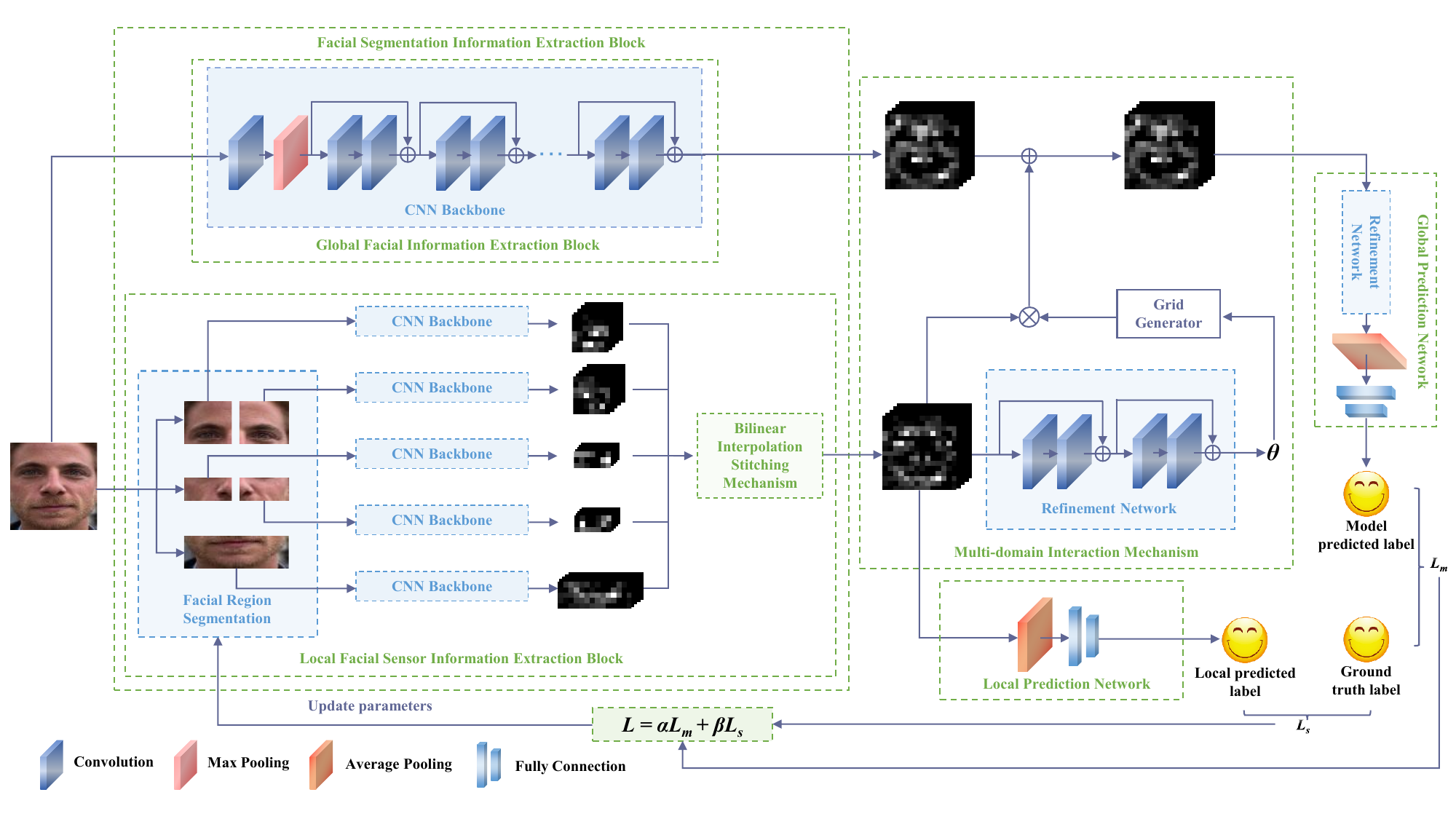}
    \caption{Network archiecture of PCNN.}
    \label{fig_structure}
\end{figure*}

\section{THE PROPOSED METHOD}
\subsection{Network Architecture}
To improve robustness of a facial recognition classifier for complex scenes, i.e. varying poses and occlusions, we use perception information to guide a discriminative learning method to achieve an efficient model, a perception CNN (PCNN) for facial expression recognition, whose network architecture can be shown in Fig. \ref{fig_structure}. PCNN includes a facial segmentation information extraction block (FSIEB), a multi-domain interaction mechanism (MDIM) and a facial semantic loss function. FSIEB mainly depends on two steps to finish facial feature extractions. The first step uses a CNN Backbone \cite{he_deep_2016} as well as global facial information extraction block (GFIEB) composed of some stacked convolutional layers, a max-pooling layer and residual learning operations to extract whole facial structural information. Due to complex scenes, i.e. varying poses and occlusions, only using a captured face image to extract facial information may lose facial emotional information within parts of sense organs to decrease performance of classification algorithms for FER. According to geometry classification algorithms, we can see that key areas based on sense organs have rich emotional information \cite{wegrzyn_mapping_2017}. Inspired by that, we use divided five regions based on sense organs to guide CNN to learn local emotional information for overcoming effects of varying poses for FER, while reasons of divided five areas can be given in the second paragraph of Section IV.C. In this paper, we respectively learn local perception information via different sense organ areas, and register obtained local emotional information to correct native effect of damaged factors, i.e. varying poses. Finally, we fuse obtained registered features and obtained normal global facial information to ensure performance of our PCNN for FER under complex scenes containing varying poses and occlusions. That is, the second step utilizes a local facial sense information extraction block (LFSIEB) to extract perception information via respectively learning sensor organ information of different local regions and a bilinear interpolation stitching mechanism \cite{gouraud_continuous_1971} to obtain facial segmentation features to improve significance of obtained emotional features for FER. Two obtained features from GFIEB and LFSIEB can be interacted via MDIM. Also, a facial semantic loss function is used to verify the effectiveness of obtained global facial information and local sense information to improve performance of obtained PCNN for FER, whose detailed information can be shown in Section III. D. The mentioned process can be formulated as follows.
\begin{equation}
    \begin{aligned}
        O_{PCNN} & = PCNN(I)                     \\
                 & = MDIM(FSIEB(I))              \\
                 & = MDIM(GFIEB(I) \& LFSIEB(I))
    \end{aligned}
\end{equation}
where $I$ and $O_{PCNN}$ denote the input image and predicted result of PCNN. $PCNN$, $FSIEB$, $MDIM$, $LFSIEB$ and $GFIEB$ denote functions of PCNN, FSIEB, MDIM, LFSIEB and GFIEB, respectively. $\&$ is a parallel operation.

\subsection{Facial Segmentation Information Extraction Block}
FSIEB is composed of two parts, i.e. GFIEB and LFSIEB. GFIEB relies on a CNN Backbone \cite{he_deep_2016} to extract global facial information. Also, LFSIEB containing three parts is responsible for learning local sensor organ information to obtain more salient information to improve the effect of our PCNN.

\begin{figure}[ht]
    \centering
    \includegraphics[width=0.6\linewidth]{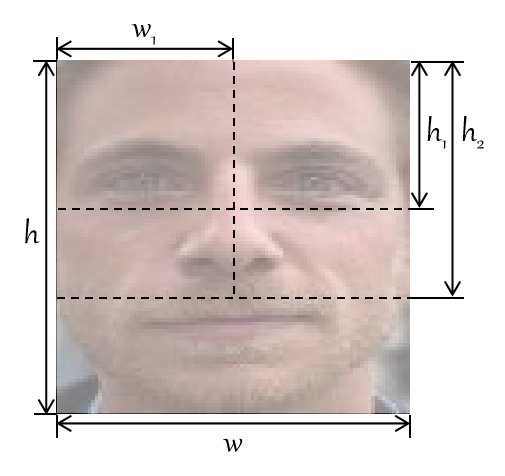}
    \caption{Sketch figures of a facial region segmentation.}
    \label{fig_seg}
\end{figure}

The first part exploits PyTorch \cite{paszke_pytorch_2019} to segment a whole facial image to obtain five sense organ areas, according to the proportion of 10:3:7 in terms of height. In Fig.\ref{fig_seg}, $h$ is set to 20,  $h_1$ is set to 10 and $h_2$ is set to 17, where $h$ is assumed as 20 to only show division principle in this paper. Width of $w$ can be cropped, according to the proportion of 1:1. That is, $w_1 = \frac{1}{2}w$. The second part exploits five parallel CNN Backbones to respectively learn local emotional information as well as perception information from five regions. The third part utilizes a bilinear interpolation stitching mechanism to splice five obtained features from five regions. According to mentioned illustrations, we can conduct Eq.\ref{eq_3} and Eq.\ref{eq_4}.
\begin{align}
    O_{GFIEB} & = GFIEB(I) = CB(I)  \label{eq_3}                            \\
    O_{GFIEB} & = LFSIEB(I) \notag                                          \\
              & = BIS\bigl(CB(S_1(I)), CB(S_2(I)), CB(S_3(I)), \label{eq_4} \\
              & \quad CB(S_4(I)), CB(S_5(I))\bigr) \notag
\end{align}
where $CB$ and $BIS$ represent functions of a CNN Backbone \cite{he_deep_2016} and a bilinear interpolation stitching, respectively. $S_i(I)$ is used to express a cropping operation of the $i$-th area of an image, which $i$ is set to [1,5]. $O_{GFIEB}$ and $O_{LFSIEB}$ are respectively outputs of GFIEB and LFSIEB, which are interacted via a multi-domain interaction mechanism in Section III. C.
\subsection{Multi-domain Interaction Mechanism}
A multi-domain interaction mechanism exploits four phases to interact obtained features from GFIEB and LFSIEB. The first phase utilizes a 4-layer residual network as a location network (also regarded as refinement network) \cite{jaderberg_spatial_2015} to compute angles of different pixel points. The second phase utilizes a grid generator \cite{jaderberg_spatial_2015} to obtain the shape of obtained features same as the output of GFIEB. The third phase applies a multiplication operation to register the obtained emotional features from the LFSIEB and the obtained features of third phase in the MDIM. First three steps are implemented via STN \cite{jaderberg_spatial_2015}. The fourth phase utilizes a residual learning operation to fuse obtained features from GFIEB and LFSIEB. Mentioned illustrations can be listed by the following equation.
\begin{equation}
    \begin{aligned}
        O_{MDIM} & = MDIM(O_{GFIEB}, O_{LFSIEB})        \\
                 & = (GG(RN(O_{LFSIEB}), O_{GFIEB})     \\
                 & \quad \times O_{LFSIEB}) + O_{GFIEB}
    \end{aligned}
\end{equation}
where $RN$ and $GG$ stand for functions of a refinement network and a grid generator, respectively. $\times$ is used to symbol a multiplication operation. $+$ is exploited to express a residual learning operation. $O_{MDIM}$ represents the output of MDIM.

\subsection{Facial Semantic Loss Function}
To ensure performance of our PCNN for FER, we design a two-phase facial semantic loss function. The first phase measures differences between predicted labels of our PCNN and given ground truth to verify classification effect of our PCNN for FER. The second phase utilizes predicted labels of LFSIEB and given ground truth to verify effectiveness of obtained local emotional features. Specifically, predicted labels of GFIEB can be obtained via a global prediction network (GPN) \cite{krizhevsky_imagenet_2012} containing a refinement network, a max pooling operation and a fully connection layer. Predicted labels of LFSIEB can be obtained via a local prediction network (LPN) \cite{krizhevsky_imagenet_2012} containing a max pooling operation and a fully connection layer. Besides, balance parameters $\alpha$ and $\beta$ are used to make a tradeoff between the two-phase loss function. The process above can be presented as follows.
\begin{equation}
    \begin{aligned}
        L & = \alpha L_{G} + \beta L_{L}                        \\
          & = \alpha CrossEntropyLoss(O_{PCNN}, GL)             \\
          & \quad + \beta CrossEntropyLoss(LPN(O_{LFSIEB}), GL) \\
          & = \alpha CrossEntropyLoss(GPN(O_{MDIM}), GL)        \\
          & \quad + \beta CrossEntropyLoss(LPN(O_{LFSIEB}), GL) \\
    \end{aligned}
\end{equation}
where $L_{G}$ and $L_{L}$ represent the global and local loss functions, respectively. $CrossEntropyLoss$ is the function of the cross-entropy loss. $LPN$ and $GPN$ are used to denote functions of LPN and GPN, respectively. $GL$ is the given ground truth label.

\section{EXPERIMENTS}
\subsection{Datasets}
To fully evaluate performance of our PCNN for FER, public datasets containing RAF-DB \cite{li_reliable_2017},  CK+\cite{lucey_extended_2010}, JAFFE \cite{lyons_coding_1998}, \cite{lyons_excavating_2021}, FER2013 \cite{goodfellow_challenges_2013}, FERPlus \cite{barsoum_training_2016} and Occlusion and Pose Variant Datasets \cite{k_wang_region_2020} are chosen to conduct comparative experiments. More detailed information of these datasets can be shown as follows.

\begin{figure}[ht]
    \centering
    \includegraphics[width=\linewidth]{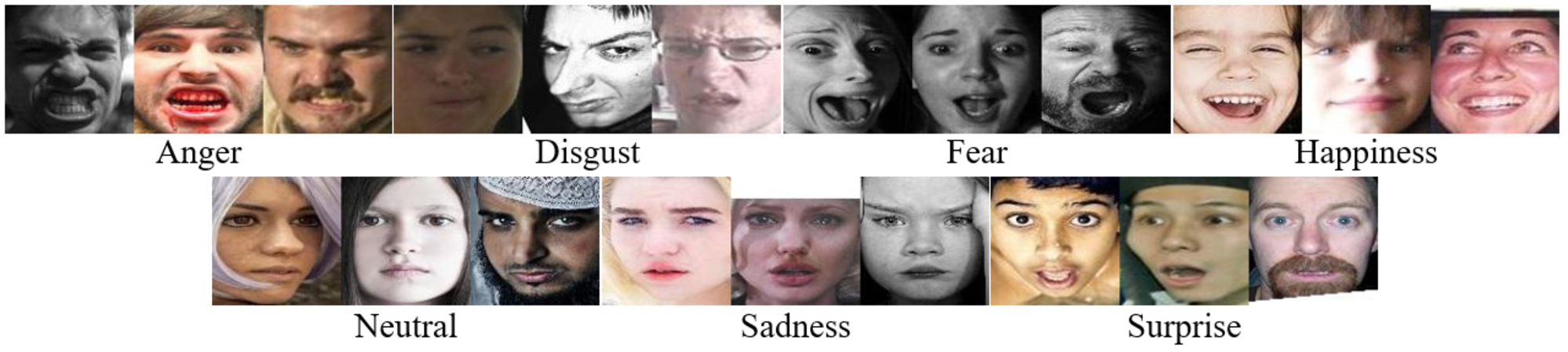}
    \caption{Seven expression facial images from the RAF-DB dataset.}
    \label{fig_rafdb}
\end{figure}

The RAF-DB dataset \cite{li_reliable_2017} includes about 30,000 facial expression images, which are captured via 315 staff members from university students and faculties to annotate these images with facial expressions. Also, six basic expressions and neutral expressions are selected from a series of expressions, including smile, terror, giggle, surprise, cry, anger, fear, shock, disgust, and so on. The seven expressions can be shown in Fig. \ref{fig_rafdb}. 12,271 images are used for training and 3,068 for testing our PCNN in this paper, where all chosen facial images are from the single expression sub-set in the RAF-DB dataset.

\begin{figure}[ht]
    \centering
    \includegraphics[width=\linewidth]{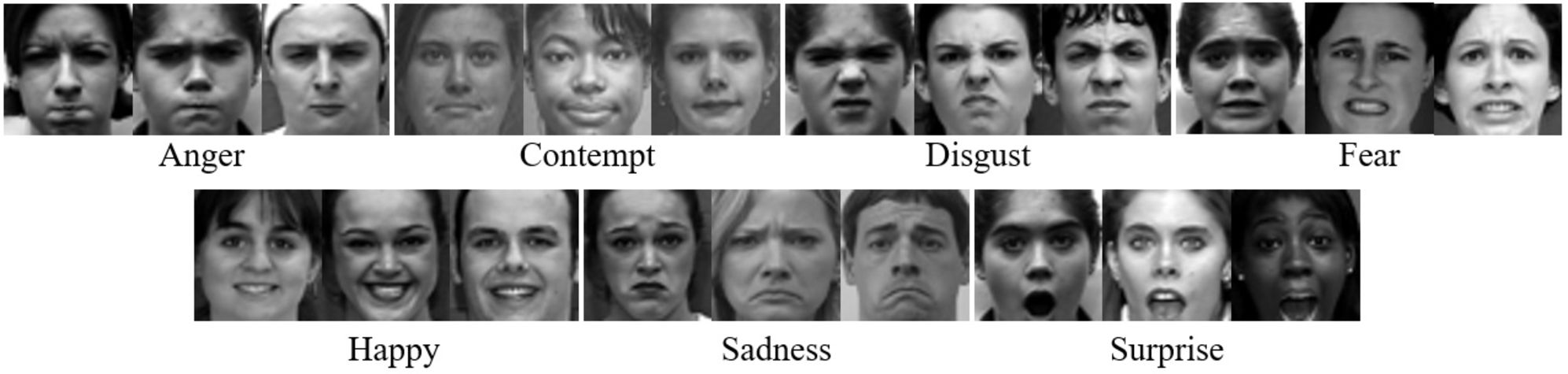}
    \caption{Seven expression facial images from the CK+ dataset.}
    \label{fig_ck}
\end{figure}

The CK+ dataset \cite{lucey_extended_2010} is a video dataset composed of 593 image sequences captured from 123 volunteers with seven expressions: surprise, anger, joy, disgust, fear, sadness and neutrality. Specifically, the neutral expression uses the first frame of each sequence as corresponding emotional expression image. Other expressions use the last frame as corresponding emotional expression image. Part of visual images with seven emotions from the CK+ dataset can be shown in Fig. \ref{fig_ck}.

\begin{figure}[ht]
    \centering
    \includegraphics[width=\linewidth]{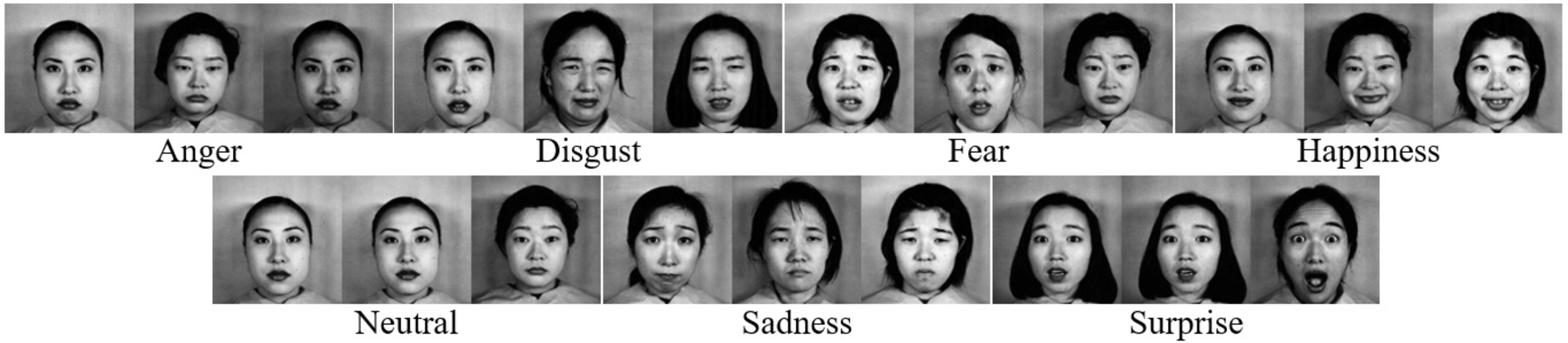}
    \caption{Seven expression facial images from the JAFFE dataset.}
    \label{fig_jaffe}
\end{figure}

The JAFFE dataset \cite{lyons_coding_1998,lyons_excavating_2021} comprises frontal facial expression images of 10 Japanese female participants. Each individual displays seven different expressions, including neutral, happy, surprise, sad, angry, fear, and disgusted. The dataset has 213 expression images, with each expression represented by 3 to 4 varied images. A notable strength of the JAFFE dataset lies on its accessible and standardized labeling of expressions. Fig. \ref{fig_jaffe} shows some visual samples of this dataset.

\begin{figure}[ht]
    \centering
    \includegraphics[width=\linewidth]{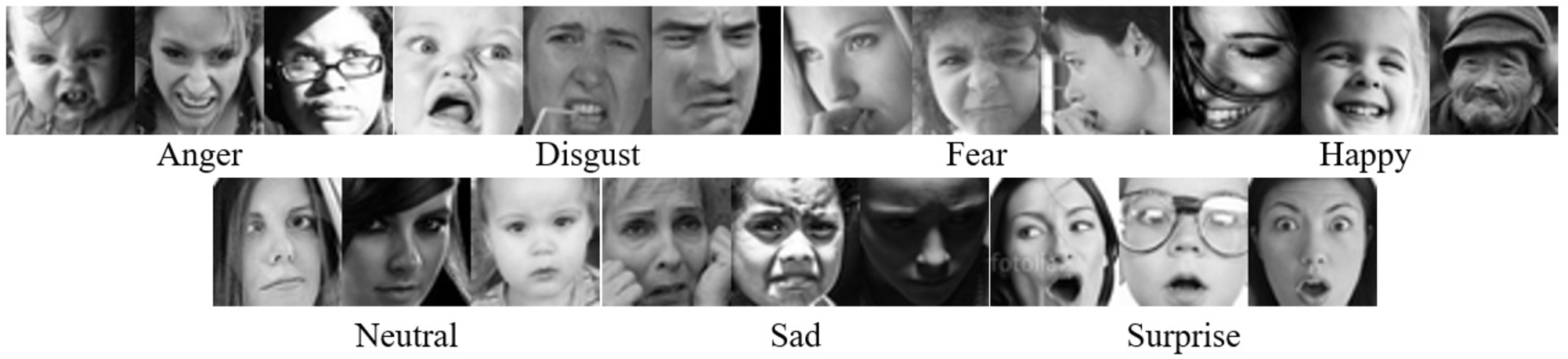}
    \caption{Several different emotional images from the FER2013 dataset.}
    \label{fig_fer2013}
\end{figure}

The FER2013 \cite{goodfellow_challenges_2013} is launched at the international machine learning conference (ICML) in 2013 containing 28,709 public training images, 3,589 public test images, which are combined as a training dataset and 3,589 private images used as a test dataset. Each image is a 48 $\times$ 48 grayscale image. Fig. \ref{fig_fer2013} shows part of visual images with different emotions.

FERPlus \cite{barsoum_training_2016} is the calibration version of FER2013, which is further cleaned and relabeled. For single label expression learning, facial images can be chosen via voting method in Ref. \cite{goodfellow_challenges_2013}. Also, NF (Not a Face) class, the unknown class and none class images are removed to achieve a cleaner result. FERPlus remains 35,714 face images in total, which contains 28559 training images and 3,573 testing images. Visual facial images of FERPlus are as follows.
\begin{figure}[ht]
    \centering
    \includegraphics[width=\linewidth]{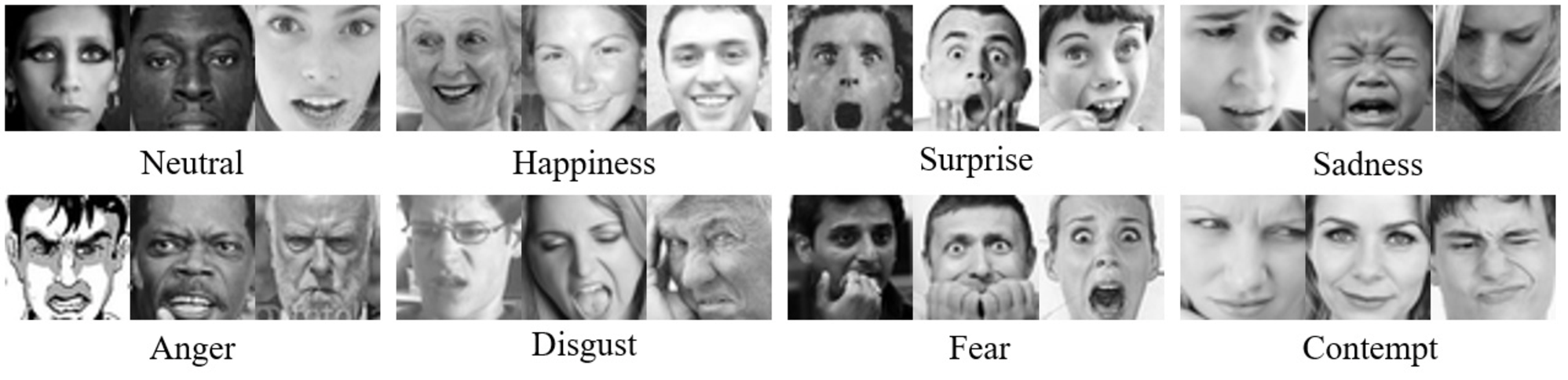}
    \caption{Visual facial images from FERPlus dataset.}
    \label{fig_ferplus}
\end{figure}

\begin{figure}[ht]
    \centering
    \includegraphics[width=\linewidth]{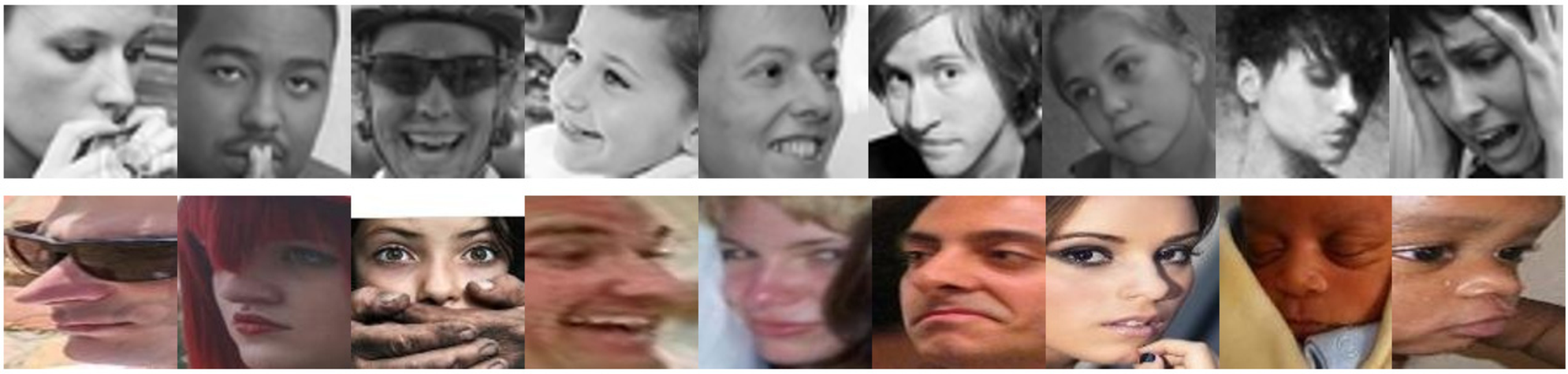}
    \caption{Visual facial images from the FERPlus dataset (Top) and RAF-DB dataset (Bottom) for occlusions and varying poses.}
    \label{fig_occlusion_pose}
\end{figure}

Occlusion and Pose Variant Datasets \cite{k_wang_region_2020} are subsets of RAF-DB and FERPlus in terms of occlusion and varying poses. Occlusions have many types, i.e. wearing a mask, wearing glasses and objects in the upper/bottom faces. Additionally, variant poses can be divided into two categories, i.e. poses larger than 30 degrees and poses larger than 45 degrees.  More visual facial images of the mentioned datasets can be seen in Fig. \ref{fig_occlusion_pose}.

\subsection{Implementation details}
Taking complexity and computational costs into account, Section III refers to CNN backbone as well as ResNet-18 \cite{he_deep_2016} in Fig. \ref{fig_structure} to make our PCNN to improve performance of FER. The CNN backbone is first pre-trained on the Ms-celeb-1m dataset \cite{guo_ms-celeb-1m_2016}. Also, our model is trained on an AMD EPYC 9554 64-Core Processor and one GPU of Nvidia GeForce GTX 4090 with Nvidia CUDA 11.7, and implemented with PyTorch 2.4.1 toolbox. All input images are reshaped to 224 $\times$ 224. The model parameters are optimized through 100 epochs training and the standard Stochastic Gradient Descent \cite{robbins_stochastic_1951} with momentum. Besides, the initial learning rate is set as 0.01, with momentum as 0.9, weight decay as 0.0001, and batch size as 256.

\subsection{Ablation Studies}
It is known that deep CNNs rely on deep networks to extract facial structural information rather than learning local facial emotional information for FER in general \cite{li_deep_2022}. Due to complex scenes containing varying poses and occlusions, the lack of robustness of these methods may result in the decrease of effectiveness in FER. In this paper, we simultaneously use structural information and facial semantics to learn more effective emotional information for improving adaptive ability of our PCNN for FER, which contains three components: a facial semantic information extraction block, a multi-domain interaction mechanism and a facial semantic loss function. FSIEB mainly extracts perception information by using sense organ information to guide CNNs to improve accuracy of FER from muscle changes. Also, it extracts global facial information by a CNN. MDIM interacts obtained facial emotional information and obtained global facial information via registration and fusion mechanisms to obtain more expressive information to improve the effect of our PCNN for FER. To prevent insufficiency of interaction between global facial information and local emotional information, a two-phase semantic loss function is designed. More detailed information of these techniques in terms of rationality and effectiveness can be illustrated as follows.

\begin{table}[ht]
    \caption{Accuracy of different methods on RAF-DB and FERPlus.}
    \label{table:cropping}
    \begin{center}
        \begin{tabular}{ccc}
            \toprule  %
            \multirow{2}{*}{Methods}        & \multicolumn{2}{c}{Accuracy(\%)}           \\  %
            \cmidrule(lr){2-3}
                                            & RAF-DB                           & FERPlus \\
            \midrule
            PCNN without cropping           & 88.11                            & 84.95   \\
            PCNN with two random cropping   & 85.95                            & 84.83   \\
            PCNN with three cropping        & 88.04                            & 84.55   \\
            PCNN with four cropping         & 87.91                            & 84.83   \\
            PCNN without balance parameters & 88.76                            & 85.06   \\
            PCNN                            & 89.31                            & 85.59   \\
            \bottomrule
        \end{tabular}
    \end{center}
\end{table}

\textbf{Facial semantic information extraction block:}  Due to differences of different application scenes, only using discriminative learning methods based on deep networks may be suffered from challenges in terms of adaptive ability of a classification model for FER via learning global structural information rather than facial attribute information to express face images for recognizing facial expressions. Inspired by that, we use facial attribute information to guide CNNs to implement interactions between structural information and facial attribute to improve robustness of an obtained PCNN for FER. Also, extracting these information can depend on a facial semantic information extraction block as well as FSIEB. FSIEB uses multi-view idea to simultaneously learn global facial information and local emotional information. That is, GFIEB utilizes a upper network as well as CNN Backbone based on front 14 layers of ResNet18 \cite{he_deep_2016} to learn global facial information. Local facial sense information extraction block as well as LFSIEB utilizes lower network based on front 14 layers of ResNet18 \cite{he_deep_2016} to learn perception information based on sense organs to obtain local emotional facial information, which is implemented via three steps. The first step of the LFSIEB exploits a facial region segmentation operation to divide a face image into five areas containing eyes and eyebrows, left and right zygomatic regions and mouth. Mentioned five segmentation areas follow the rules as follows.  1) Divided areas should keep integrity of sense organs in each area. 2) Important sense organs should be divided into different areas. According to mentioned rules, we conduct several experiments to verify optimal choice of mentioned five areas. Firstly, we directly put a face image as an input of a lower network rather than dividing this face image to learn facial information. Learned information from an upper and lower networks are global facial information, which ignores local emotional information of learning important sense organs. Its effect may be inferior to a mentioned method based five areas for facial recognition expression. In TABLE \ref{table:cropping}, `PCNN without balance parameters' has better accuracy than that of `PCNN without cropping' on RAF-DB and FERPlus, where `PCNN without cropping' represents PCNN of an input of a lower network without segmentation areas. Also, `PCNN without balance parameters' is PCNN with $\alpha = 10$ and $\beta = 10$, which more introductions of two parameters can be given in Section III.D. Secondly, we randomly crop a face image to two parts as inputs of lower networks to learn local emotional information for FER. However, it may break the first rule above to break integrity of part sense organs to decrease performance of FER. That is, accuracy of `PCNN with two random cropping' is lower than `PCNN without balance parameters' on RAF-DB and FERPlus for FER in TABLE \ref{table:cropping}, where `PCNN with two random cropping' expresses the PCNN model with with two random cropping areas as inputs of the two lower networks. Because `PCNN without cropping' does not break integrity of sense organs, it has higher accuracy than that of `PCNN with two random cropping' on RAF-DB and FERPlus for FER in TABLE \ref{table:cropping}. Thirdly, we horizontally cut on an input face of local facial twice as inputs of FESIEB as well as `PCNN with three cropping' for FER to again verify the first rule mentioned above, which may break integrity of left and right zygomatic regions to decrease performance of obtained PCNN for FER. As shown in TABLE \ref{table:cropping}, we can see that `PCNN with three cropping' has lower accuracy in contrast to `PCNN without balance parameters' in FER. Finally, to verify the first and second rules, we vertically and horizontally evenly cut a face image once to obtain four areas as inputs of four CNN Backbones as well as `PCNN with four cropping' in the LFSIEB, which breaks integrity of mouth in an area to decrease performance of our PCNN for FER. That can be verified via `PCNN with four cropping' and `PCNN without balance parameters' in TABLE \ref{table:cropping}. According to mentioned analysis and verification, we can see that mentioned methods based on five cropping areas in the LFSIEB is optimal choice in this paper. Specifically, these CNN Backbones are pre-trained on Ms-celeb-1m dataset \cite{guo_ms-celeb-1m_2016} and then fine-tuned it to guarantee good performance of our PCNN in FER. The second step of the LFSIEB utilizes five parallel CNN Backbones to extract local emotional information. The third step of the LFSIEB exploits a bilinear interpolation stitching mechanism \cite{gouraud_continuous_1971} to fuse obtained features from the second step. Effectiveness of GFIEB can be tested via comparing `GFIEB' with a 4-layer residual classifier and `ResNet18' as well as a method without GFIEB and LFSIEB in TABLE \ref{table:ablation}, which ResNet18 is not pretrained on Ms-celeb-1m dataset \cite{guo_ms-celeb-1m_2016}. That is, our proposed GFIEB has improvements of 43.80\% on RAF-DB and 11.82\% on FERPLus than that of raw ResNet18 for FER. Effectiveness of LFSIEB can be verified via `a combination of GFIEB and LFSIEB' with a 4-layer residual classifier and `GFIEB' with a 4-layer residual classifier in TABLE \ref{table:cropping}. Also, effectiveness of FSIEB can be proved via comparing `GFIEB and LFSIEB' and `ResNet18' in TABLE \ref{table:cropping}.

\begin{table}[ht]
    \caption{Accuracy of different techniques in a PCNN on RAF-DB and FERPlus.}
    \label{table:ablation}
    \begin{center}
        \resizebox{\linewidth}{!}{
            \begin{tabular}{ccccccc}
                \toprule
                \multicolumn{2}{c}{FSIEB} & \multirow{2}{*}{MDIM} & \multirow{2}{*}{CBP} & \multirow{2}{*}{BP} & \multirow{2}{*}{RAF-DB (\%)} & \multirow{2}{*}{FERPlus (\%)}         \\ 
                \cmidrule(lr){1-2}
                GFIEB                     & LFSIEB                &                      &                     &                         &                          &       \\
                \midrule 
                                          &                       &                      &                     &                         & 44.20                    & 73.15 \\
                \checkmark                &                       &                      &                     &                         & 88.00                    & 84.97 \\
                \checkmark                & \checkmark            &                      &                     &                         & 88.14                    & 85.03 \\
                \checkmark                & \checkmark            & \checkmark           &                     &                         & 88.64                    & 85.11 \\
                \checkmark                & \checkmark            & \checkmark           & \checkmark          &                         & 88.76                    & 85.06 \\
                \checkmark                & \checkmark            & \checkmark           &                     & \checkmark              & 89.31                    & 85.59 \\
                \bottomrule
            \end{tabular}
        }
    \end{center}
\end{table}

\textbf{Multi-domain Interaction Mechanism:} FSIEB has extracted global facial features and local emotional features and a muti-domain interaction mechanism as well as MDIM is used to interact these obtained information. MDIM uses four phases to maintain good performance for FER. The first phase utilizes last four layers of ResNet18 as well as a refinement network to compute angles of different pixel points. The second phase utilizes a grid generator \cite{jaderberg_spatial_2015} to determine shape of obtained local emotional features, which is the same as obtained features of GFIEB. The third phase exploits a multiplication operation to register obtained emotional features from the LFSIEB and obtained global features from the second phase to obtain full emotional features. Front three phases in the multi-domain interaction mechanism based on STN \cite{jaderberg_spatial_2015} are used to register obtained local emotional information based on sense organs to obtain salient expressions for FER. Thus, parameters of mentioned front three phases are learnable rather than static according to different scenes in the training process. The fourth phase applies a residual learning operation to fuse obtained global facial features of GFIEB and local emotional features of LFSIEB. Effectiveness of MDIM can be verified via a combination model of GFIEB, LFSIEB and MDIM, and the model with single FSIEB containing GFIEB and LFSIEB on RAF-DB and FERPlus in TABLE \ref{table:ablation}.

\textbf{Facial Semantic Loss Function:} To fully ensure interaction effects of global facial information and local emotional information, two-phase facial semantic loss function is employed. The first phase measures difference between predicted result of PCNN and given ground truth to test effect of our PCNN for FER. The second phase measures differences between predicted result of LFSIEB and given ground truth. Also, predicted labels of LFSIEB and PCNN can be introduced in Section III. D. Two phases can reach a tradeoff via two parameters of $\alpha$ and $\beta$. Effectiveness of two-phase loss can be proved via comparing our PCNN and PCNN containing FSIEB, MDIM and BP, and a combination of FSIEB and MDIM in TABLE \ref{table:ablation}, where BP denotes balanced parameters. Choices of two parameters are shown as follows. The first step sets $\alpha = 10$ and $\beta = 10$, where their effectiveness can be tested via a combination of FSIEB, MDIM and CBP, a combination of FSIEB, MDIM and BP on RAF-DB and FERPlus in TABLE \ref{table:ablation}, where CBP stands for certain balanced parameters of $\alpha$ and $\beta$. To monitor loss function values of training process, the second step enlarges ten times to each balanced parameter and keeps fixed sum of two balanced parameters. To keep consistence with traditional machine learning methods for normalizing operation, the third step chooses even to set two parameters. According to previous geometry based recognition algorithms for FER, we can see that facial key areas include rich expression information, which is important for FER \cite{tian_recognizing_2001}. The four step breaks the rule of  $\frac{\beta}{\alpha} > 0.5$.

\begin{figure}[ht]
    \centering
    \includegraphics[width=0.8\linewidth]{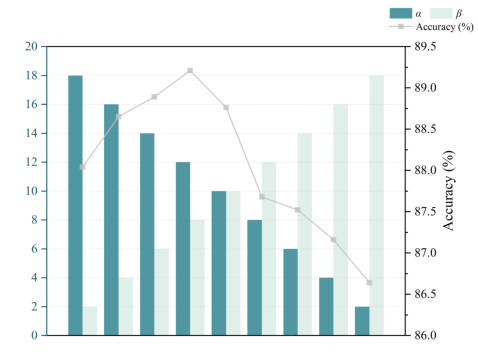}
    \caption{Comparison in different hyper-parameters}
    \label{fig_alpha_beta}
\end{figure}

Thus, $\alpha = 12$ and $\beta = 8$ are set as balanced parameters in this paper. Their effectiveness can be shown in Fig. \ref{fig_alpha_beta}, where they have obtained the highest accuracy with $\alpha$ of 12 and $\beta$ of 8 for FER. Thus, this setting of balanced parameters is reasonable. According to mentioned analysis and verification, we can see that the proposed facial semantic information extraction block, multi-domain interaction mechanism and facial semantic loss function are reasonable and useful for FER in this paper.

\subsection{Experimental Results}
In this section, we compare the proposed PCNN with several deep learning-based methods, i.e., DLP-CNN \cite{li_reliable_2017}, gACNN \cite{li_occlusion_2019}, pACNN \cite{li_occlusion_2019}, SCAN \cite{gera_landmark_2021}, Compact CNN \cite{kuo_compact_2018}, FDRL \cite{ruan_feature_2021}, SCN \cite{wang_suppressing_2020}, RAN \cite{k_wang_region_2020}, RUL \cite{zhang_relative_2021}, EfficientFace \cite{zhao_robust_2021}, Poster \cite{zheng_poster_2023}, BReG-Net-50 \cite{hasani_breg-next_2022}, ResNet-18 \cite{he_deep_2016}, BReG-NeXt-50 \cite{hasani_breg-next_2022}, ALAW \cite{xie_adaptive_2021}, PASM \cite{liu_point_2021}, TFE-JL \cite{li_facial_2018}, VGG13+MV \cite{barsoum_training_2016}, VGG13+CEL \cite{barsoum_training_2016}, VGG13+PLD \cite{barsoum_training_2016}, DICNN \cite{saurav_dual_2022}, LEWEL \cite{huang_learning_2022}, PCL \cite{liu_pose-disentangled_2023}, IPA2LT \cite{zeng_facial_2018}, CERN \cite{li_decoding_2025}, DENet \cite{li_unconstrained_2023}, PCARNet \cite{qi_novel_2024}, MA-Net \cite{zhao_learning_2021}, RARN \cite{li_learning_2022}, DMUE (ResBet-18) \cite{she_dive_2021}, AMP-Net \cite{liu_adaptive_2022}, FER-VT \cite{huang_facial_2021}, PACVT \cite{liu_patch_2023}, FERMixNet \cite{huang_fermixnet_2024} on RAF-DB, CK+, JAFFE and FER-2013 facial expression datasets. These datasets are conducted for lab scenes containing CK+ \cite{lucey_extended_2010}, JAFFE \cite{lyons_coding_1998,lyons_excavating_2021}, real scenes containing FER-2013 \cite{goodfellow_challenges_2013}, FERPlus\cite{barsoum_training_2016} and RAF-DB \cite{li_reliable_2017} and complex scenes of varying poses and occlusions, i.e., Occlusion and Pose Variant Datasets \cite{k_wang_region_2020}. Due to small dataset of JAFFE and less experimental results of different methods on JAFFE for FER, to keep fair, we use results of authoritative journals on JAFFE \cite{wasi_grefel_2024} as comparative results to test performance of our model.

For FER under lab scenes, TABLE \ref{table:ck+} summarizes the performance comparison of different methods on the CK+ dataset. Our PCNN achieves a recognition accuracy of 100.00\%, surpassing all compared methods. Specifically, PCNN surpasses DLPCNN \cite{li_reliable_2017}, which enhances the discriminative power of deep features by preserving locality closeness while maximizing inter-class scatters by 4.22\%. Compared to the attention-based gACNN \cite{li_occlusion_2019} and pACNN \cite{li_occlusion_2019}, our model achieves advantages of 3.60\% and 2.97\% for FER, respectively. It also exceeds the spatial-channel attention network SCAN \cite{gera_landmark_2021}, which leverages local-global attention without landmark guidance by 2.93\%. Furthermore, PCNN outperforms Compact CNN \cite{kuo_compact_2018} and FDRL \cite{ruan_feature_2021} by 1.53\% and 0.46\%, respectively. To verify robustness of our PCNN for lab scenes, we conduct experiments on JAFFE. As shown in the TABLE \ref{table:jaffe}, our PCNN achieves the highest facial recognition accuracy of 96.49\%. More precisely, our PCNN obtains a performance gain of 10.16\% over SCN \cite{wang_suppressing_2020}. Compared with RAN \cite{k_wang_region_2020}, RUL \cite{zhang_relative_2021} and EfficientFace \cite{zhao_robust_2021}, our PCNN boosts the performance on facial expression recognition by 7.82\%, 4.16\% and 4.16\%, respectively. For the sub-optimal Poster \cite{zheng_poster_2023}, our PCNN has a significant increase of 1.92\%. These results indicate that the proposed PCNN achieves competitive performance on lab scene.

\begin{table}[ht]
    \caption{Performance comparison on CK+.}
    \label{table:ck+}
    \begin{center}
        \begin{tabular}{lc}
            \toprule
            \textbf{Methods}                    & \textbf{Accuracy(\%)} \\
            \midrule
            DLP-CNN \cite{li_reliable_2017}     & 95.78                 \\
            gACNN \cite{li_occlusion_2019}      & 96.40                 \\
            pACNN \cite{li_occlusion_2019}      & 97.03                 \\
            SCAN \cite{gera_landmark_2021}      & 97.07                 \\
            Compact CNN \cite{kuo_compact_2018} & 98.47                 \\
            FDRL \cite{ruan_feature_2021}       & 99.54                 \\
            PCNN(Ours)                      & 100.00                \\
            \bottomrule
        \end{tabular}
    \end{center}
\end{table}

\begin{table}[ht]
    \caption{Performance comparison on JAFFE.}
    \label{table:jaffe}
    \begin{center}
        \begin{tabular}{lc}
            \toprule
            \textbf{Methods}                      & \textbf{Accuracy(\%)} \\
            \midrule
            SCN \cite{wang_suppressing_2020}      & 86.33                 \\
            RAN \cite{k_wang_region_2020}         & 88.67                 \\
            RUL \cite{zhang_relative_2021}        & 92.33                 \\
            EfficientFace \cite{zhao_robust_2021} & 92.33                 \\
            Poster \cite{zheng_poster_2023}       & 94.57                 \\
            PCNN(Ours)                        & 96.49                 \\
            \bottomrule
        \end{tabular}
    \end{center}
\end{table}

\begin{figure}[htbp]
    \centering
    \subfloat[FER2013 \label{fig_fer2013_cfm}]{%
        \includegraphics[width=0.5\linewidth]{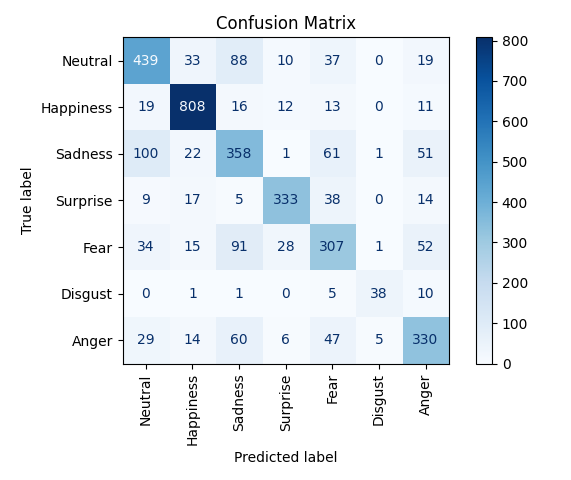}%
    }
    \hfill
    \subfloat[RAF-DB \label{fig_rafdb_cfm}]{%
        \includegraphics[width=0.5\linewidth]{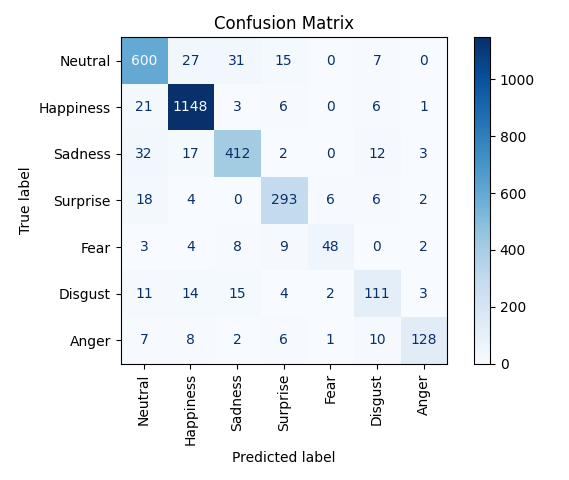}%
    }
    \caption{Confusion matrix of our PCNN on different datasets containing FER2013 and RAF-DB.}
    \label{fig_cfm}
\end{figure}

For FER of our PCNN under the real scenes, we conduct comparative experiments on the FER-2013, FERPlus and RAF-DB. FER-2013 exists error labels, which may affect effects of different methods for FER. From TABLE \ref{table:fer2013}, our PCNN has achieved relatively excellent facial expression recognition results compared with all comparative methods. In contrast to both BReG-Net-50 \cite{hasani_breg-next_2022} and ResNet18 \cite{he_deep_2016}, our proposed PCNN achieves a recognition accuracy of 72.80\%, demonstrating a significant enhancement in performance. Similarly, our PCNN outperforms BReG-NeXt-50 \cite{hasani_breg-next_2022} and ALAW \cite{xie_adaptive_2021}. Although the recognition accuracy of PCNN lags behind only 0.79\% compared to the best-performing PASM \cite{liu_point_2021}. According to mentioned comparisons, we can see that our PCNN is effective for FER-2013. To overcome drawback of FER-2013, FERPlus are conducted via relabeling FER-2013 for FER. As shown in TABLE \ref{table:ferplus}, our PCNN achieves an accuracy of 85.59\% for FER, which surpasses TFE-JL \cite{li_facial_2018} by 1.30\%, and also exceeds the VGG13-based models \cite{barsoum_training_2016}. Specifically, our PCNN outperforms VGG13+MV (majority voting) by 1.11\%, VGG13+CEL (cross-entropy loss) by 0.63\%, and VGG13+PLD (probabilistic label drawing) by 0.24\% for FER. Additionally, PCNN shows a slight improvement of 0.30\% over DICNN \cite{saurav_dual_2022}. These results demonstrate the competitive performance of our method on the FERPlus dataset. For color facial expression recognition, RAF-DB is used to conduct comparative experiments. TABLE \ref{table:rafdb} presents the quantitative comparison on the RAF-DB dataset between our PCNN and existing methods. Compared to PCARNet \cite{qi_novel_2024}, which combines a pyramid convolution module with an enhanced convolution attention mechanism to effectively extract expression features, our PCNN achieves a 1.79\% gain for FER. Furthermore, our model surpasses DMUE (ResNet-18) \cite{she_dive_2021}, a method integrating both pose features and facial expression descriptors by 0.55\%. Compared with MA-Net \cite{zhao_learning_2021}, our method achieves a 0.91\% improvement, further validating the powerful capability of PCNN in complex facial expression recognition tasks. According to the confusion matrix as shown in Fig. \ref{fig_cfm}, PCNN demonstrated enhanced discriminative capability for expression-adjacent categories. On RAF-DB, which contains sufficient samples from major categories and rich color information, the model achieved high recognition accuracy for primary emotions such as neutral, happy, sad, and surprise, indicating its ability to learn stable global and local expression-related features from complex backgrounds and color variations. Furthermore, even in FER-2013, which is characterized by higher levels of noise and uneven data distribution, PCNN maintains strong performance in recognizing more distinguishable categories such as happiness and surprise, illustrating a degree of robustness. Although some misclassifications occurred among fine-grained negative expressions with subtle inter-class differences (e.g., fear, sad, and certain instances of anger or neutral), PCNN still delivers competitive results in these challenging categories. This suggests its advantage in enhancing fine-grained facial expression representation and suppressing intra-class variability. In summary, PCNN exhibits promising baseline capabilities in addressing issues related to expression similarity and imbalanced data distribution. According to mentioned comparisons, we can see that our PCNN is suitable to FER under the real scenes.

\begin{table}[ht]
    \caption{Performance comparison on FER-2013.}
    \label{table:fer2013}
    \begin{center}
        \begin{tabular}{lc}
            \toprule
            \textbf{Methods}                          & \textbf{Accuracy(\%)} \\
            \midrule
            BReG-Net-50 \cite{hasani_breg-next_2022}  & 69.21                 \\
            ResNet-18 \cite{he_deep_2016}             & 72.3                  \\
            BReG-NeXt-50 \cite{hasani_breg-next_2022} & 71.53                 \\
            ALAW \cite{xie_adaptive_2021}             & 72.67                 \\
            PASM \cite{liu_point_2021}                & 73.59                 \\
            PCNN(Ours)                            & 72.80                 \\
            \bottomrule
        \end{tabular}
    \end{center}
\end{table}

\begin{table}[ht]
    \caption{Performance comparison on FERPlus.}
    \label{table:ferplus}
    \begin{center}
        \begin{tabular}{lc}
            \toprule
            \textbf{Methods}                       & \textbf{Accuracy(\%)} \\
            \midrule
            TFE-JL \cite{li_facial_2018}           & 84.29                 \\
            VGG13+MV \cite{barsoum_training_2016}  & 84.48                 \\
            VGG13+CEL \cite{barsoum_training_2016} & 84.96                 \\
            VGG13+PLD \cite{barsoum_training_2016} & 85.35                 \\
            DICNN \cite{saurav_dual_2022}          & 85.29                 \\
            PCNN(Ours)                         & 85.59                 \\
            \bottomrule
        \end{tabular}
    \end{center}
\end{table}

\begin{table}[ht]
    \caption{Performance comparison on RAF-DB.}
    \label{table:rafdb}
    \begin{center}
        \begin{tabular}{lc}
            \toprule
            \textbf{Methods}                      & \textbf{Accuracy(\%)} \\
            \midrule
            IPA2LT \cite{zeng_facial_2018}        & 86.77                 \\
            CERN \cite{li_decoding_2025}          & 86.82                 \\
            RAN \cite{k_wang_region_2020}         & 86.90                 \\
            DENet \cite{li_unconstrained_2023}    & 87.35                 \\
            PCARNet \cite{qi_novel_2024}          & 87.52                 \\
            SCN \cite{wang_suppressing_2020}      & 88.14                 \\
            MA-Net \cite{zhao_learning_2021}      & 88.40                 \\
            EfficientFace \cite{zhao_robust_2021} & 88.36                 \\
            RARN \cite{li_learning_2022}          & 88.72                 \\
            DMUE (ResBet-18) \cite{she_dive_2021} & 88.76                 \\
            AMP-Net \cite{liu_adaptive_2022}      & 89.25                 \\
            PCNN(Ours)                            & 89.31                 \\
            \bottomrule
        \end{tabular}
    \end{center}
\end{table}

\begin{figure}
    \centering
    \includegraphics[width=0.8\linewidth]{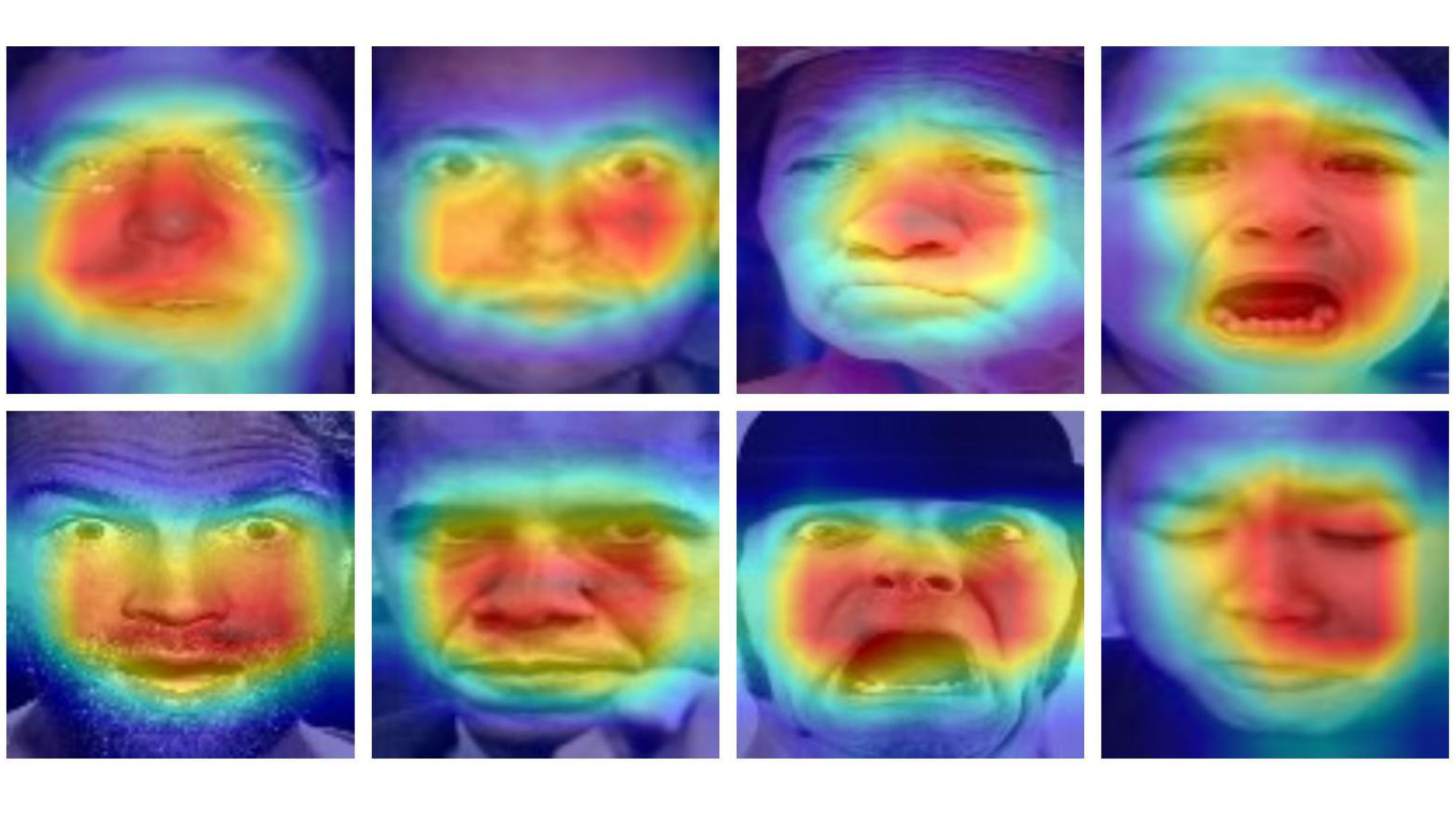}
    \caption{Visualization results of our PCNN by Gradcam++ \cite{chattopadhay_grad-cam_2018} on RAF-DB dataset}
    \label{fig_visual}
\end{figure}

To intuitively demonstrate capability of PCNN in extracting emotion-related information from facial images, we employed Grad-CAM++ to visualize the feature maps derived from the refinement network of the GPN in a PCNN model trained on the RAF-DB dataset. The results are shown in Figure \ref{fig_visual}. By integrating the ESIEB and MDIM modules with a facial semantic loss function, PCNN effectively preserves global semantic context while sensitively capturing subtle local variations in regions such as the peri-ocular area, brow region, mouth corners, and cheeks. The Grad-CAM++ \cite{chattopadhay_grad-cam_2018} visualizations reveal that PCNN consistently attends to facial regions semantically relevant to expression cues—for instance, the mouth and cheeks in happiness, or the eyelids and forehead in surprise. Through structured local feature learning and registration-based fusion, PCNN is capable of effectively extracting and leveraging facial emotional clues, thereby achieving robust facial expression recognition.

For FER under complex scenes, i.e. occlusion and varying poses, we evaluate our PCNN on the Occlusion and Pose Variant Datasets for FER \cite{k_wang_region_2020} including Occlusion-RAFDB, Occlusion-FERPlus, Pose-RAFDB, and Pose-FERPlus. On the two occlusion datasets, as shown in TABLE \ref{table:occlusion}, our PCNN respectively achieves accuracy of 85.56\% and 84.63\%  on Occlusion-RAFDB and Occlusion-FERPlus, which are competitive with existing methods, showing a 1.11\% improvement over RAN \cite{k_wang_region_2020} on Occlusion-RAFDB and slightly surpassing PACVT \cite{liu_patch_2023} by 0.33\% on Occlusion-FERPlus. Mentioned analysis demonstrates the robustness of PCNN in handling occluded facial expression images. On the two pose variation datasets, as shown in TABLE \ref{table:pose}, our PCNN achieves competitive performance across different methods. On the Pose-RAFDB, it respectively attains accuracy of 89.74\% and 87.28\%  under face poses of V30 and V45, outperforming RAN \cite{k_wang_region_2020} by 3.00\% and 2.08\%, and showing a slight improvement over FERMixNet \cite{huang_fermixnet_2024}. On the Pose-FERPlus, PCNN reaches accuracy of 88.38\% (V30) and 87.52\% (V45), which has improvements of 6.15\% and 7.12\% than that of RAN and remains comparable to FERMixNet. These results indicate that PCNN maintains robust recognition capability under varying poses and occlusions.

\begin{table}[htbp]
    \caption{Performance comparison on Occlusion datasets.}
    \label{table:occlusion}
    \begin{center}
        \resizebox{\linewidth}{!}{
            \begin{tabular}{lcc}
                \toprule
                \multirow{2}{*}{Methods} & \multicolumn{2}{c}{Accuracy on occlusion datasets (\%)} \\
                & \multicolumn{1}{c}{Occlusion-RAFDB} & \multicolumn{1}{c}{Occlusion-FERPlus} \\
                \midrule 
                ResNet-18 \cite{he_deep_2016}      & 80.19                    & 73.33                      \\
                RAN \cite{k_wang_region_2020}      & 82.72                    & 83.63                      \\
                MA-Net \cite{zhao_learning_2021}   & 83.65                    & -                          \\
                FER-VT \cite{huang_facial_2021}    & 84.32                    & -                          \\
                DENet \cite{li_unconstrained_2023} & 85.44                    & -                          \\
                PACVT \cite{liu_patch_2023}        & 83.59                    & 84.74                      \\
                PCNN(Ours)                         & 85.56                    & 84.63                      \\
                \bottomrule
            \end{tabular}
        }
    \end{center}
\end{table}

\begin{table}[ht]
    \caption{Performance comparison on pose change datasets.}
    \label{table:pose}
    \begin{center}
        \begin{tabular}{lcc} 
            \toprule
            \multirow{2}{*}{Methods}                    & \multicolumn{2}{c}{Accuracy on pose change datasets (\%)} \\
            & \multicolumn{1}{c}{Pose-RAFDB V30} & \multicolumn{1}{c}{Pose-RAFDB V45} \\
            \midrule
            ResNet-18 \cite{he_deep_2016}         & 78.11        & 75.50        \\
            RAN \cite{k_wang_region_2020}         & 86.74        & 85.20        \\
            DENet \cite{li_unconstrained_2023}    & 87.57        & 86.92        \\
            FERMixNet \cite{huang_fermixnet_2024} & 89.01        & 87.63        \\
            PCNN(Ours)                        & 89.74        & 87.28        \\
            \midrule
            Methods & \multicolumn{1}{c}{Pose-FERPlus V30} & \multicolumn{1}{c}{Pose-FERPlus V45} \\
            \midrule
            ResNet-18 \cite{he_deep_2016}         & 78.11        & 75.50        \\
            RAN \cite{k_wang_region_2020}         & 82.23        & 80.40        \\
            FERMixNet \cite{huang_fermixnet_2024} & 88.55        & 86.73        \\
            PCNN(Ours)                            & 88.38        & 87.52        \\
            \bottomrule
        \end{tabular}
    \end{center}
\end{table}

To test practicality of our PCNN for mobile devices, i.e. phones and cameras, we verify complexities of our PCNN and different methods for FER. As shown in Table \ref{table:complexity}, we compare the computational efficiency of PCNN with several existing methods in terms of parameter count, running time, and computational complexity. PCNN contains 51.00 million parameters, which is fewer than AMP-Net \cite{liu_adaptive_2022} (59.44M), RARN \cite{li_learning_2022} (70.37M), and MA-Net \cite{zhao_learning_2021} (63.54M). In terms of inference speed, PCNN processes each sample in $5.83\pm0.04$ ms, corresponding to 171.53 items per second. This inference time is shorter than that of RARN \cite{li_learning_2022} (10.96 ms), MA-Net \cite{zhao_learning_2021} (9.48 ms), and AMP-Net \cite{liu_adaptive_2022} (13.96 ms). Although PCNN requires 14.80 GMACs, which is higher than the other methods (ranging from 3.60 to 4.48 GMACs), our PCNN is superior in parameter count, running time and throughput compared to the other methods. Thus, our PCNN is competitive between model complexity and inference efficiency, making it suitable for practical deployment scenarios.
Additionally, compared with SOTA methods, i.e., EAC \cite{zhang_learn_2022}, APViT \cite{xue_vision_2022}, S2D \cite{chen_static_2024}, TL \cite{akhand_facial_2021} and POSTER \cite{zheng_poster_2023} for FER on RAF-DB, our PCNN is slightly inferior to them. For example,
EAC achieve the accuracy of 90.35\%, outperforming our PCNN by 1.04\% on RAF-DB dataset. APViT gains the accuracy of 91.98\%, which is 2.67\% higher than our PCNN. Although our PCNN is weak in comparsion to these SOTA methods for FER, novelty of our method is that simultaneously facial semantic information guides a CNN to obtain more robust emotional information for FER, which is differing from these methods based on improving network architectures. That may bring an inspiration for scholars. Also, our methods are simple to be easily implemented and configured. In summary, our methods are competitive for FER.

\begin{table}[ht]
    \caption{Complexities and running time of different methods for FER.}
    \label{table:complexity}
    \begin{center}
        \resizebox{\linewidth}{!}{
            \begin{tabular}{lcccc}
                \toprule
                \textbf{Methods}                 & \textbf{Parameters (M)} & \textbf{Running time (ms)} & \textbf{MACs (G)} & \textbf{Items/s} \\
                \midrule
                PCNN (Ours)                      & 51.00                   & $5.83\pm0.04$              & 14.80             & 171.53           \\
                AMP-Net \cite{liu_adaptive_2022} & 59.44                   & $13.96\pm0.42$             & 4.48              & 71.63            \\
                RARN \cite{li_learning_2022}     & 70.37                   & $10.96\pm0.22$             & 3.60              & 91.22            \\
                MA-Net \cite{zhao_learning_2021} & 63.54                   & $9.48\pm0.07$              & 3.66              & 105.48           \\
                \bottomrule
            \end{tabular}
        }
    \end{center}
\end{table}

\section{Conclusion}
This paper presents a perception CNN via using facial segmentation based on sensor organ information and facial information to recognize facial expressions. Firstly, it uses facial semantic information extraction block to respectively learn global facial information and local emotional information. Secondly, it utilizes a multi-domain interaction mechanism to register and fuse obtained global facial information and local emotional information to improve the effect of our PCNN for facial expression recognition. Finally, we a facial semantic loss to guarantee effectiveness of interacting obtained global facial information and local emotional information and verify performance of our proposed PCNN. Experimental results prove the effect of our PCNN for FER in different scenes.

\bibliographystyle{IEEEtran} 
\bibliography{ref} 

\begin{IEEEbiography}[{\includegraphics[width=1in,height=1.25in,clip,keepaspectratio]{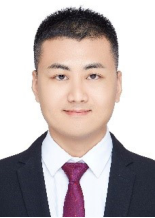}}]{Chunwei Tian}(Member, IEEE) received his Ph.D. degree from Harbin Institute of Technology in 2021. He is a professor with the School of Computer Science and Technology, Harbin Institute of Technology, Harbin, China. His interests include image restoration, image generation, intelligent transportation, multimodal large model. He has published over 100 scientific papers ininternational journals and conferences containing IEEE TIP, IEEE TNNLS, IEEE TMM, IEEE TCSVT, IEEE TSMC, IEEE TIV, IEEE TCE, IEEE TIM, PATTERN RECOGNITION, INFORMATION FUSION, NEURAL NETWORKS, ACM MM, NeurIPS, etc. He has 8 highly cited papers, 5 cover papers of NN and TMM, a distinction prize paper of PR, an excellent paper of CAAI Transactions on Intelligence Technology, etc. He has serviced an editor/GE of CAAI Transactions on Intelligence Technology, Dence Technology, IEEE TCE, etc.\end{IEEEbiography}
\begin{IEEEbiography}[{\includegraphics[width=1in,height=1.25in,clip,keepaspectratio]{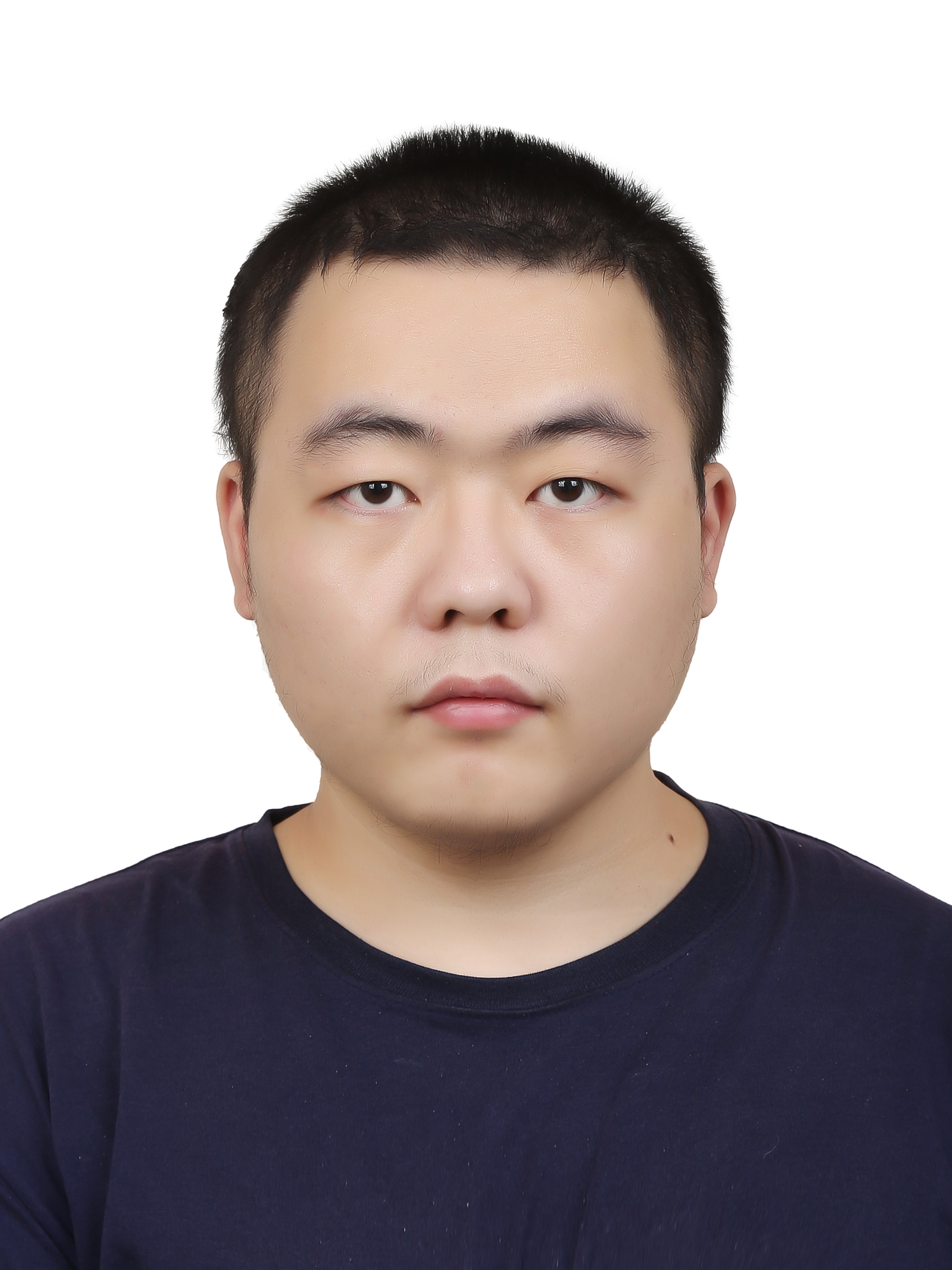}}]{Jingyuan Xie} received his B.S. degree in the School of Software, Northwestern Polytechnical University, Xi'an, China. He is currently working toward the M.S. degree in the School of Software, Northwestern Polytechnical University. His research interests include image recognition and generation.\end{IEEEbiography}
\begin{IEEEbiography}[{\includegraphics[width=1in,height=1.25in,clip,keepaspectratio]{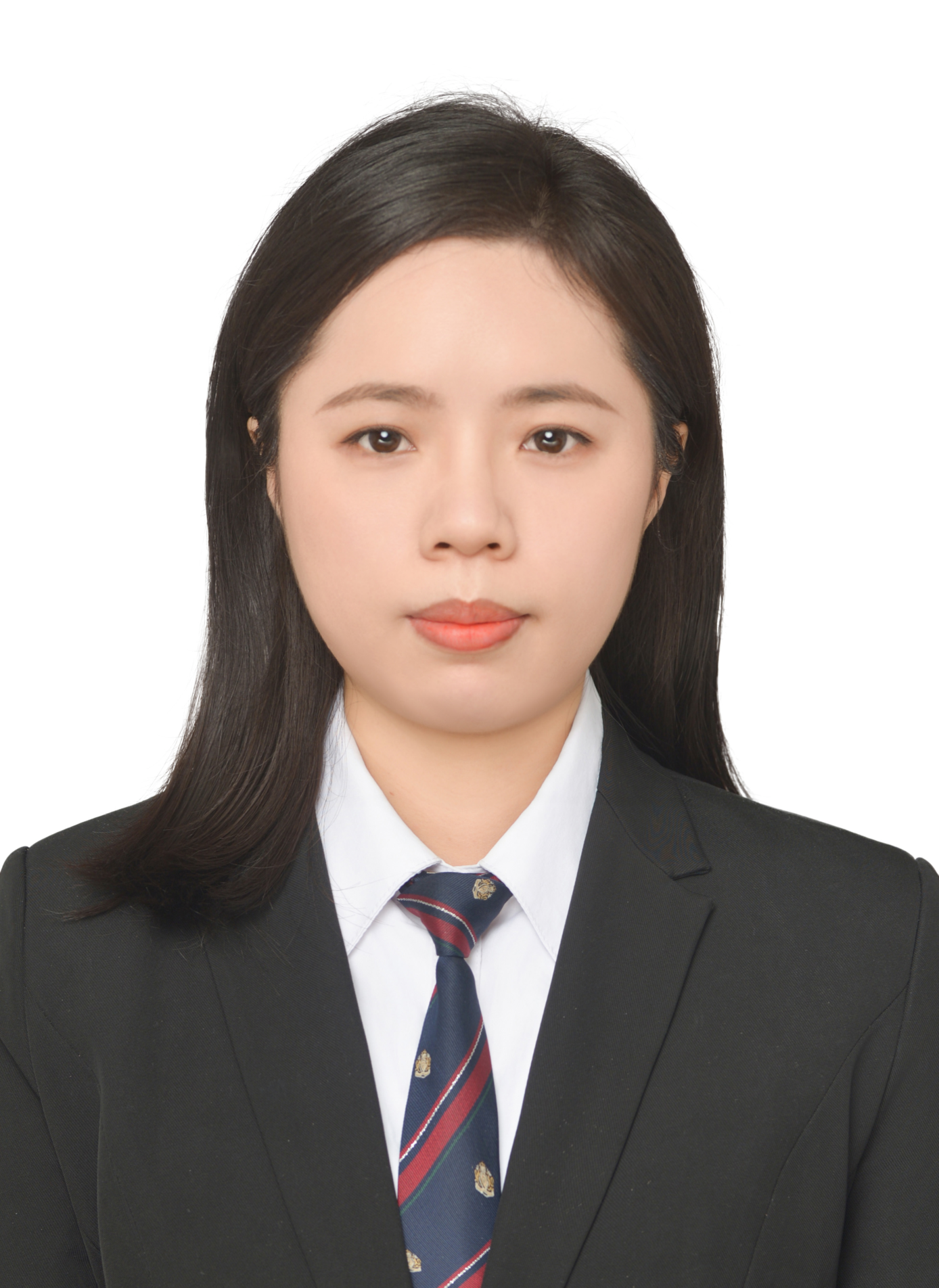}}]{Lingjun Li} received the M.S. degree from the Shaanxi Normal University, Xi'an, China, in 2018, and the Ph.D. degree from Northwestern Polytechnical University, Xi’an, China, in 2024. She is currently working at Zhengzhou University of Light Industry, Zhengzhou, China. Her research interests include computer vision and remote sensing image processing, especially on object detection and scene classification.\end{IEEEbiography}
\begin{IEEEbiography}[{\includegraphics[width=1in,height=1.25in,clip,keepaspectratio]{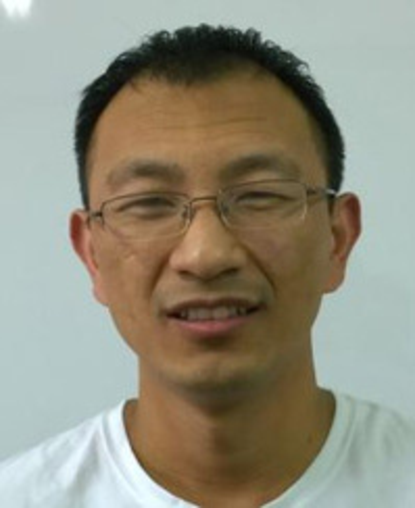}}]{Wangmeng Zuo}(Senior Member, IEEE) received the PhD degree in computer application technology from the Harbin Institute of Technology, Harbin, China, in 2007. He is currently a professor with the School of Computer Science and Technology, Harbin Institute of Technology. He has published more than 100 papers in top-tier academic journals and conferences. His current research interests include image enhancement and restoration, image and face editing, and visual generation. He has served as an associate editor for IEEE Transactions on Pattern Analysis and Machine Intelligence and IEEE Transactions on Image Processing.\end{IEEEbiography}
\begin{IEEEbiography}[{\includegraphics[width=1in,height=1.25in,clip,keepaspectratio]{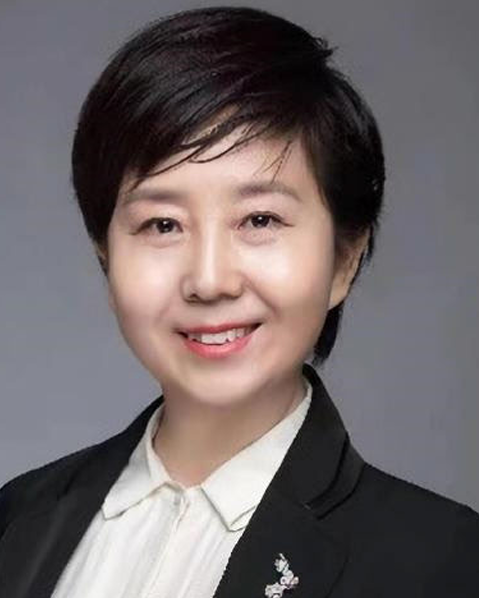}}]{Yanning Zhang}(Fellow, IEEE) received the B.S. degree from the Dalian University of Science and Engineering, Dalian, China, in 1988, and the M.S.and Ph.D. degrees from Northwestern Polytechnical University, Xi’an, China, in 1993 and 1996, respectively. She is currently Professor with the School of Computer Science, Northwestern Polytechnical University. She is also the Organization Chair of the Ninth Asian Conference on Computer Vision 2009. She has authored/coauthored more than 200 papers in international journals, conferences, and Chinese key journals. Her research interests include signal and image processing, computer vision, and pattern recognition.\end{IEEEbiography}
\begin{IEEEbiography}[{\includegraphics[width=1in,height=1.25in,clip,keepaspectratio]{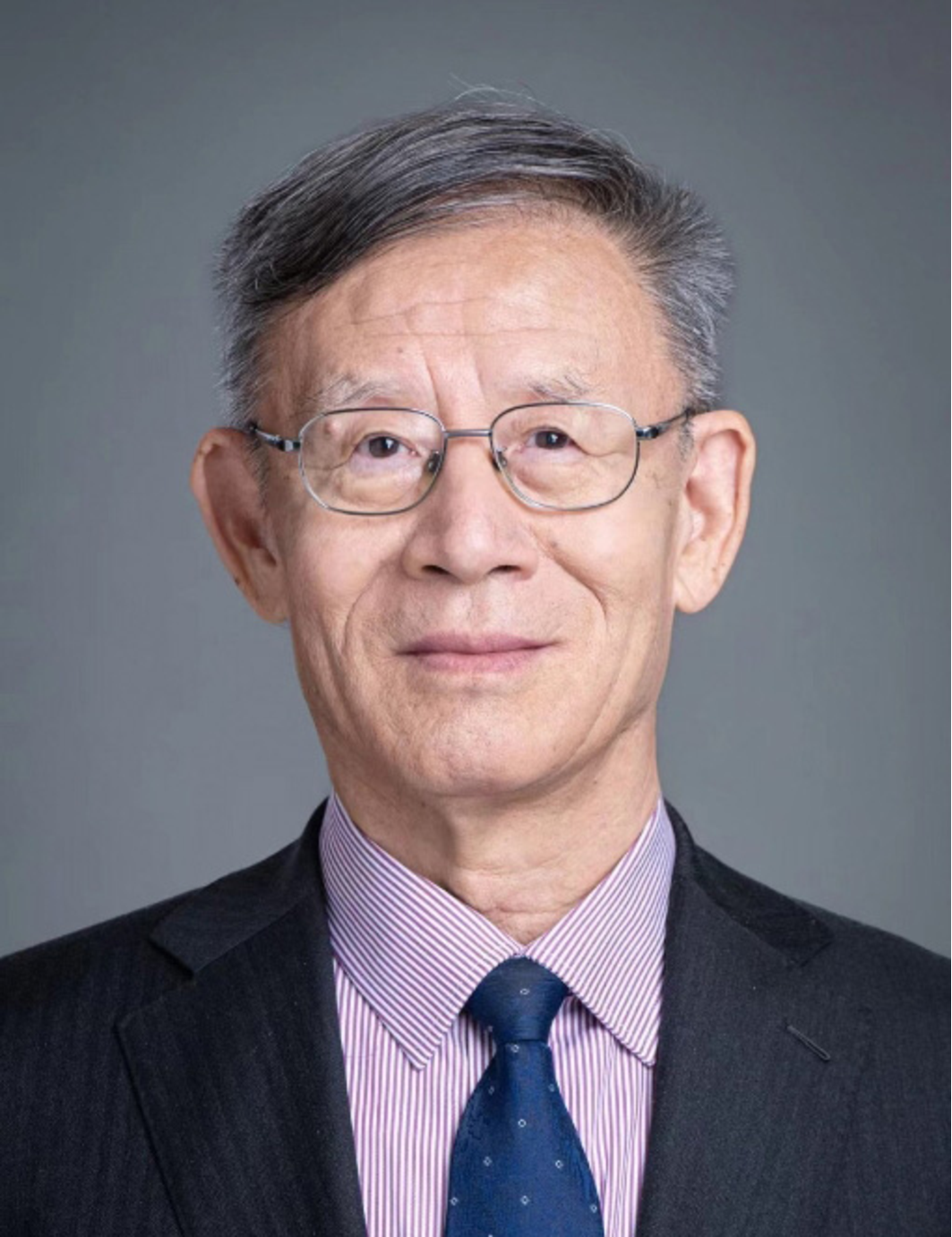}}]{David Zhang}(Life Fellow, IEEE) received the Ph.D. degree from Peking University, Beijing, China, in 1974, the M.S. and first Ph.D. degrees in computer science from Harbin Institute of Technology, Harbin, China, in 1982 and 1985, respectively,and the second Ph.D. degree in electrical and computer engineering from the University of Waterloo, Waterloo, ON, Canada, in 1994. From 1986 to 1988, he was a Post-Doctoral Fellow with Tsinghua University, Beijing, and then an Associate Professor with Academia Sinica, Beijing. He has been a Chair Professor with Hong Kong Polytechnic University, Hong Kong, where he is the Founding Director of the Biometrics Research Centre (UGC/CRC) supported by Hong Kong SAR Government since 1998. He is currently a Distinguished Presidential Chair Professor with The Chinese University of Hong Kong (Shenzhen), Shenzhen, China. Over the past 40 years, he has been working on pattern recognition, image processing, and biometrics, where many research results have been awarded and somecreated directions, including medical biometrics and computerized TCM, are famous in the world. He has published more than 20 monographs, more than 500 international journal articles, and more than 50 patents from USA, Japan, and China. For 8 years, he has been continuously listed as a Global Highly Cited Researcher in Engineering by Clarivate Analytics. He is also ranked 70th with H-index 133 at Top 1000 Scientists for International Computer Science in 2023. Prof. Zhang has been selected as a fellow of both Royal Society of Canada (RSC) and the Canadian Academy of Engineering (CAE). He is also a Croucher Senior Research Fellow, a Distinguished Speaker of the IEEE Computer Society, and an IAPR and an AAIA Fellow.\end{IEEEbiography}

\vfill

\end{document}